\documentclass[10pt, a4paper]{article}
\usepackage{lrec}
\usepackage{multibib}
\newcites{languageresource}{Language Resources}
\usepackage{graphicx}
\usepackage{tabularx}
\usepackage{soul}

\usepackage{epstopdf}
\usepackage{arabtex}
\usepackage[utf8]{inputenc}
\usepackage{adjustbox}
\usepackage[T1]{fontenc}

\usepackage{hyperref}
\usepackage{xstring}

\usepackage{color}

\setcode{utf8}
 \setarab

 \arabtrue

\title{Automated Prediction of Medieval Arabic Diacritics}

\name{Khalid Alnajjar, Mika Hämäläinen, Niko Partanen, Jack Rueter}

\address{Department of computer science, Department of digital humanities, \\ Department of Finnish, Finno-Ugrian and Scandinavian Studies  \\
         University of Helsinki, Finland \\
         \{firstname.lastname\}@helsinki.fi\\}

\abstract{
This study uses a character level neural machine translation approach trained on a long short-term memory-based bi-directional recurrent neural network architecture for diacritization of Medieval Arabic. The results improve from the online tool used as a baseline. A diacritization model have been published openly through an easy to use Python package available on PyPi and Zenodo. 
We have found that context size should be considered when optimizing a feasible prediction model. \\ \newline  }

\begin{document}

\maketitleabstract

\section{Introduction}

The Arabic language is customarily written without marking the diacritic characters that indicate vowels, their absence, long consonants etc. This results in numerous possible readings for individual word forms, which is a major challenge, for instance, in natural language processing. Additionally, this ambiguity poses difficulties for the non-native learners of the language who might not be aware of the correct pronunciation or lexeme in question when reading text.

Having diacritics in place is beneficial for any endeavors in processing speech computationally. For text-to-speech applications, the missing diacritics and the phonemes they represent have to be predicted one way or another before vocalizing the text as synthesized speech. Furthermore, the inverted process of converting speech to text can benefit from data that has diacritics in place as it brings the textual representation closer to phonetic realization. 

Although diacritization of Modern Standard Arabic is a topic that has received quite some attention in the academic research, much less research has been conducted on historical Arabic. The dictionary data our research is based on is not linguistically very different from Modern Standard Arabic, but as it nevertheless represents an older variety, the differences are particularly pronounced in different lexicon and genre. Although Medieval Arabic and Modern Standard Arabic have a lot in common, capturing all the nuances with tools tailored for the modern language variant is not possible without tools that have been built specifically for historical data.  As Medieval Arabic texts are becoming increasingly available in digital format, due to numerous projects involving their preservation and digitization, there is an increasing need for NLP applications that work within this domain as well.

Furthermore, diacritization of Arabic text reduces the level of ambiguity of the tokens. This is useful for such applied NLP tasks where a high degree of ambiguity is an issue. For instance when studying word usage in historical texts any kind of noise or ambiguity in the data is undesirable. In a wider context, work on Arabic diacritization will also benefit other languages that have similar, largely ambiguous orthographies, such as Hebrew or Aramaic. Besides adding missing vowel characters, the task at hand is similar to adding other comparable information, i.e. historically motivated circumflexes in French or orthographically unmarked weight signs in Russian.

From this point of view, the Arabic diacritization can be described as a task to recover linguistic information that is not fully encoded in the character-level orthographic representation. There are systematically different possible readings for individual word forms, and the correct reading must be disambiguated by using the surrounding context. How challenging this task ends up being is very language specific, but it can be concluded that with sufficient linguistic context the task must be possible, as the native speakers are able to perform it successfully in their daily lives.

The best working model presented in this paper has been released in an easy to use Python library\footnote{https://github.com/mikahama/haracat} \citelanguageresource{alnajjar_khalid_2020_3677375} in order to facilitate the reuse of the language resource. This also makes it easier to compare other approaches directly with our model.

\section{Related Work}

%
\begin{figure*}[!htb]
\center{\includegraphics[width=15cm]
{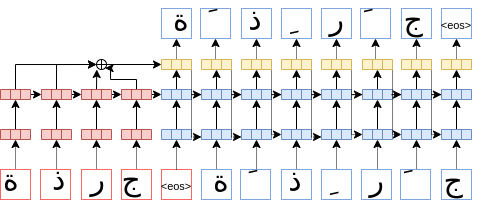}}
\caption{\label{fig:network} A high level illustration of the character-level NMT architecture}
\end{figure*}

Early statistical approaches employ a hidden Markov model for restoration of Arabic diacritics \cite{10.3115/1118637.1118641}. They train their model on the Quran. They show that using a bigram model can improve over an unigram baseline.

In the past, one take on adding diacritics to Arabic text was to use WFSTs (weighted finite state transducers) \cite{nelken2005arabic}. They combine FSTs of several different levels of abstraction: word level, letter level, and simple morphology. This line of work presents the rule-based tradition that was strong in the past in the field of NLP.

Maximum-entropy models have been used to predict diacritics in the Arabic Treebank data \cite{zitouni2009arabic}. They model the problem as a sequence classification task. They use three different types of features lexical, segment-based and part-of-speech (POS) features for the model.

LSTM networks have been used to predict diacritics on modern text \cite{abandah2015automatic}, they approach the problem as a sequence labeling task. In their approach, they train the model on a word level to predict the corresponding diacritized word forms. One limitation with this approach is that the model is limited to only operating with the exact words that were present in the training data. A similar word level LSTM has also been proposed by \cite{belinkov2015arabic}.

SVMs (support vector machines) have also been used relatively recently for predicting Arabic diacritics \cite{darwish2017arabic}. Their system is divided into two components, the first of which adds diacritics to the core word and the second of which adds diacritics to the morphological ending.

Apart from focusing on a historical language variant, our work is different from the existing contemporary LSTM based approaches by two key ways. First, we tackle this problem on the character level, which makes it possible for the model to learn to generalize diacritics even for words not present in the vocabulary. Secondly, we train our models to fully diacrtizise the input text. This means not only to predict the diacritics when absolutely needed for ambiguity resolution, but to always predict them.

Diacritization as a task is in many ways comparable to the normalization of historical or dialect texts, although there are also numerous differences. Our work with Medieval Arabic locates to the context of NLP with historical language, 
although our goal is to retrieve the original diacritization, instead of i.e. generating the corresponding Modern Standard Arabic equivalents. Similarly spoken representations of vowels often differ significantly from these variants, prediction of which is an entirely different task. This, naturally, leaves many open possibilities for future research.

%


\section{Data and Preprocessing}

For the purpose of diacritic prediction, we use the online version of the  Medieval Arabic dictionary Al-Qāmūs l-Mu\d{h}ī\d{t}: \citelanguageresource{arabic-dict}. The dictionary contains text fields (henceforth we refer to these as sentences) for each lemma, and all of these sentences have all the diacritics in place, even in the unambiguous cases. The dictionary is distributed by alphabet in the Clarin Virtual Language Observatory infrastructure\footnote{https://vlo.clarin.eu/} \citelanguageresource{goosen2014virtual} in XML format. We download dictionaries for every alphabet and extract the sentences.

We clear all the extracted sentences from any diacritics with a regular expression. This results in a parallel data set of sentences with and without diacritics. The data consists of over 58,000 sentences with a sentence length of 8 words, on the average. We shuffle and split the data so that 70\% is used for training, 15\% for validation and 15\% for evaluation.

\begin{figure}[!htb]
\center{\includegraphics[width=7cm]
{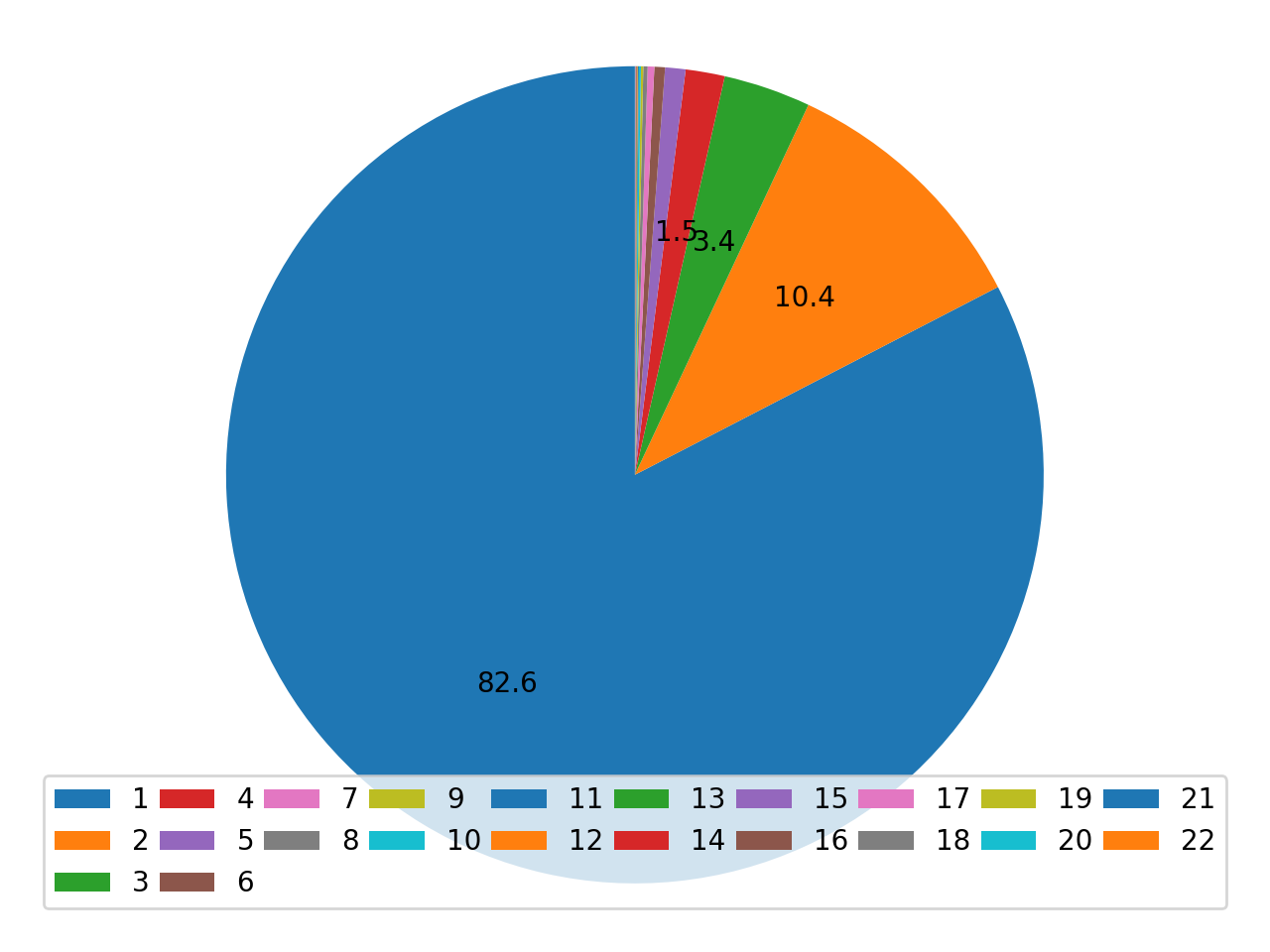}}
\caption{\label{fig:ambi} Degree of ambiguity in the corpus on type level}
\end{figure}

Figure \ref{fig:ambi} shows the degree of ambiguity of the tokens without diacritics within the corpus. This means that 82.6\% of the words (types) have only one single possibility for their diacritized variant. The maximum number of different diacritizations a word can get is 22, but this occurs only for two words in the corpus. Thus this problem is mostly a one-to-one mapping problem for most of the words.

\begin{figure}[!htb]
\center{\includegraphics[width=7cm]
{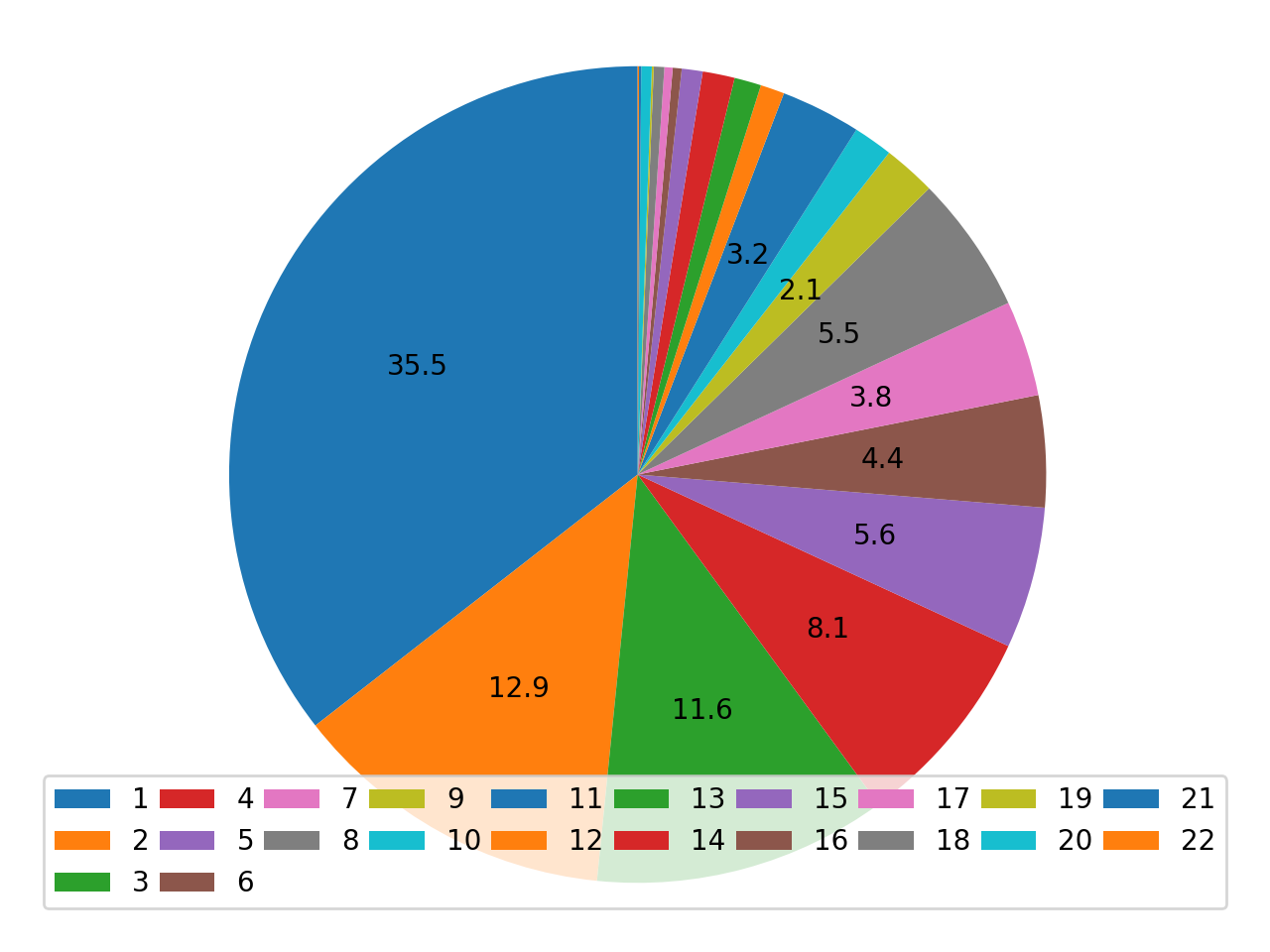}}
\caption{\label{fig:ambi2} Degree of ambiguity in the corpus on token level}
\end{figure}

However, Figure \ref{fig:ambi2} shows that on the token level, there is more ambiguity, as only 35.5\% of the tokens belong to the category of only one possible diacritization. It is more frequent on the token level to have multiple possible alternatives. This indicates that ambiguity resolution becomes important for the most frequent words.

\section{Predicting Diacritics}

\begin{table*}[!ht]
\centering
\small
\begin{tabular}{|l|c|c|c|c|c|c|c|c|c|c|c|c|c|}
\hline
model & \begin{tabular}[c]{@{}c@{}}No \\ prediction\end{tabular} & baseline & \begin{tabular}[c]{@{}c@{}}1\\ word\end{tabular} & \begin{tabular}[c]{@{}c@{}}2\\ words\end{tabular} & \begin{tabular}[c]{@{}c@{}}3\\ words\end{tabular} & \begin{tabular}[c]{@{}c@{}}4\\ words\end{tabular} & \textbf{\begin{tabular}[c]{@{}c@{}}5\\ words\end{tabular}} & \begin{tabular}[c]{@{}c@{}}6\\ words\end{tabular} & \begin{tabular}[c]{@{}c@{}}7\\ words\end{tabular} & \begin{tabular}[c]{@{}c@{}}8\\ words\end{tabular} & \begin{tabular}[c]{@{}c@{}}9\\ words\end{tabular} & \begin{tabular}[c]{@{}c@{}}10\\ words\end{tabular} & \begin{tabular}[c]{@{}c@{}}whole \\ sentence\end{tabular} \\ \hline
WER   & 87.98                                                    & 61.20    & 42.63                                            & 37.24                                             & 35.11                                             & 34.40                                             & \textbf{34.09}                                             & 36.30                                             & 39.08                                             & 41.90                                             & 46.59                                             & 49.20                                              & 58.72                                                     \\ \hline
\end{tabular}
\caption{Results of the models}
\label{tab:results}
\end{table*}

As character-level neural machine translation (NMT) models have achieved good results before in a variety of similar tasks such as spelling normalization \cite{adouane2019normalising,partanen2019dialect} and OCR post-correction \cite{hamalainen2019paft}, we have decided to opt for a character-based approach on the task of diacritic prediction as well. Character level modeling makes it possible for the model to learn diacritization even for words that have not appeared in the training data as the unit of operation is a character rather than an entire word.

We train an NMT model using a long short-term memory (LSTM) based  bi-directional recurrent neural network (BRNN) architecture on a character level for modeling the problem. As opposed to a regular one-directional RNN, the BRNN model can benefit from both left and right contexts when predicting the diacritics, as the encoding takes place from  both the beginning and end of the sequence. The architecture consists of two encoding layers and two decoding layers and the general global attention model \cite{luong2015effective}. The network is illustrated in Figure \ref{fig:network}.

We train the models by using the OpenNMT Python package \cite{opennmt} by otherwise resorting to the default settings for training that are predefined in the package. This also means that the training is done for 100,000 steps.

We experiment with multiple scenarios in terms of the length of the input sequence. We train models for predicting diacritics one word at a time, multiple words at a time and entire sentences. We experiment with chunk sizes from 1 to 10 tokens at a time. In this way, we attempt to find the optimal length of the sequence for predicting the diacritics. The longer the sequence, the more context it gives, on the one hand, and, on the other, long sequences tend to make the model perform worse \cite{partanen2019dialect}. We train all the models with the same random seed and same division between training, validation and testing data in order to make their intercomparison meaningful.

When training the model, the words are split into characters that are separated by white spaces. This also means that diacritics appear as separate tokens on the target side. Word boundaries in the models that are trained with more than one word at a time, multi-word elements, are marked with underscores (\_).
\novocalize
\begin{table}
\resizebox{\columnwidth}{!}{%
\begin{tabular}{|c|r|}
\hline
Input & \begin{tabular}[c]{@{}l@{}}\begin{RLtext}wkAn Al.s.hAbT ytmAz.hwn .ht_A \end{RLtext} \\ \begin{RLtext}ytbAd.hwn bAlb.ty_h, f'i_dA .hzbhm\end{RLtext}\\  \begin{RLtext}'mr kAnwA hm AlrjAl '.s.hAb Al'mr \end{RLtext}\end{tabular} \\ \hline
Prediction & \begin{tabular}[c]{@{}l@{}}\begin{RLtext}wkAna Al.sxxa.hAbaTu yatamaAza.huwna .ht_A \end{RLtext}\\ \begin{RLtext}yatbAda.hwna bAlbi.txxiy_h, f'i_dA .hazabahum"\end{RLtext} \\ \begin{RLtext}'m"ruN kAnwA humu AlrxxijAlu '.s".hAbu Al"'am"ri \end{RLtext} \end{tabular} \\ \hline
Correct & \begin{tabular}[c]{@{}l@{}}\begin{RLtext}wkAn Al.s.hAbTu ytmAza.hwna .ht_A \end{RLtext}\\ \begin{RLtext}yatbAda.hwna bAlbi.txxiy_hi, f'i_dA .hazabahum" \end{RLtext}\\ \begin{RLtext}'m"ruN kAnwA humu AlrxxijAla '.s.hAba Al"'am"ri\end{RLtext}\end{tabular} \\ \hline
\end{tabular}%
}
\caption{An example of the input, output and gold data}
\label{tab:example}

\end{table}

\section{Results \& Evaluation}

In this section we report and discuss the results of the models. As a baseline, we use a popular online tool\footnote{http://www.tahadz.com/mishkal} for adding diacritics to Arabic script. We report the word error rates\footnote{WER is counted with an open source tool https://github.com/nsmartinez/WERpp} (WER) for each model. The lower the word error rate, the better the model is at predicting the diacritics. WER is a common metric derived from Levenshtein edit distance \cite{Levenshtein66Binary}, and it takes into account the number of deletions $D$, substitutions $S$, insertions $I$ and the number of correct words $C$. It is calculated per sentence with the following formula:

\begin{equation}
WER = \frac{S + D + I}{S + D + C}
\end{equation}

It is important to note that we are not counting diacritic error rate, as such an evaluation metric can easily give overly positive looking results as even partially correctly predicted diacritics for a word would lower the error rate. When focusing on WER, only word with fully correctly predicted diacritics will be considered as correct, any partially wrong predictions will lead to the entire word as counted erroneous. In this respect our evaluation method differs from commonly used metrics such as DER (Diacritic Error Rate), but in our opinion WER gives the most realistic picture about the accuracy. In downstream tasks where the model could be used it would customarily be very important that complete words are diacritized correctly.

Table \ref{tab:results} shows the results of the models. The first column named \textit{No prediction} shows the WER when no diacritics are added to the words at all. This means comparing sentences without diacritics to the ones with diacritics. All of our NMT models outperform the baseline system. Table \ref{tab:example} shows an example sentence with diacritics predicted by the five word character level model.

The results suggest that having five words at a time when predicting the diacritics is optimal for the model. It gives enough contextual cues without making the sequences overly long. The results get progressively worse by removing or adding more words. The best performing character based method reduces the WER by 53.89.

Although the accuracy reached still leaves space for further experimentation, the reduction in WER is still very large and considerable. Further work is certainly needed to improve the model, and, for instance, better handling of errorenous predictions would be useful, but already the work presented here improves significantly from the previous baseline.

\section{Conclusions and Future Work}

In this paper we have approached the challenging problem of dealing with a Medieval language form of Arabic in terms of its full diacritization. Our character based NMT approach can also deliver diacritization for words not found in the in the vocabulary, as the model learns mappings from characters rather than words. Needless to say, it would be interesting in the future to attempt combining word- and character-level methods.

We have released the model trained on chunks of five words as a Python library. We have put special care in making the library as easy to use as possible so that it is not just a research resource that is difficult to install, but a library that is easy to plugin to any Python based code with next to no prior knowledge on NLP required. The library can be installed by using pip\footnote{pip install haracat}, the defacto tool for managing Python libraries, since it is distributed on PyPi with full versioning and releases uploaded automatically to Zenodo for permanent storage. As we have used only openly available resources and infrastructure, it is easy to extend to work with further resources and experiments.

An important finding that has not been studied to a satisfactory degree in the literature is the effect the number of words has on the results. With shorter chunks of words, the model does not see enough context to reach a feasible amount of accurate predictions, while with longer chunks the model sees too much context and loses its meaningfulness to the prediction. The optimal number of words for the performance seems to be five. In the future, however, it would be interesting to see if this is specific to the NMT architecture used in this paper or specific to the data set the model was trained on, or whether it is a tendency observable across architectures and data sets. 

In this vein, there is also the possibility that different tasks are just linguistically difficult to process in different degrees, and in this context various results can contain important information about linguistic complexity. As these are tasks a human agent with sufficient linguistic knowledge is able to perform, the conditions under which neural networks are able to learn similar processing may tell about contextual window necessary for various tasks.

\section{Bibliographical References}\label{reference}

\bibliographystyle{lrec}
\bibliography{lrec2020W-xample-kc}

\section{Language Resource References}
\label{lr:ref}
\bibliographystylelanguageresource{lrec}
\bibliographylanguageresource{languageresource}

\end{document}